\documentclass[sigconf]{acmart}
\AtBeginDocument{%
  }

\usepackage{fontawesome}

\usepackage{multirow} 

\usepackage{mathtools}

\usepackage{graphicx}
\usepackage{xspace}
\usepackage{colortbl}
\usepackage{xcolor}
\usepackage{tablefootnote}
\usepackage{url}

\usepackage{bm}
\usepackage{enumitem}
\usepackage{bbding}
\usepackage{hyperref}

\usepackage{array}         % 表格格式控制
\usepackage{booktabs}      % 更好看的表格线
\usepackage{makecell}      % 支持换行
\usepackage{tabularx}      % 自动适应宽度
\usepackage{caption}       % 控制标题样式

\usepackage{amsmath}
\usepackage{amssymb}
\usepackage{amsmath}

\usepackage{subfig}
\usepackage{xspace}
\usepackage{adjustbox}

\usepackage[linesnumbered,ruled,vlined]{algorithm2e}

\newcommand{\upgood}{\color{green!70!black}$\Uparrow$}
\newcommand{\downgood}{\color{red!70!black}{$\Downarrow$}}

\newcolumntype{o}{>{\columncolor{red!5}}c}
\newcolumntype{b}{>{\columncolor{blue!5}}c}
\newcolumntype{g}{>{\columncolor{lightgray!30}}c}

\newcolumntype{i}{>{\columncolor{red!15}}c}
\newcolumntype{n}{>{\columncolor{blue!15}}c}

\usepackage{multirow, hhline}

\usepackage{algorithmicx}
\usepackage{algpseudocode}
\usepackage{seqsplit}

\newcommand{\sota}{state-of-the-art\xspace}
\newcommand{\myparagraph}[1]{\vspace{1ex}\noindent\textbf{#1.}\hspace{1em}}

\newcommand{\ourmodel}{\textsf{ExeFuse}\xspace}
\newcommand{\icewswiki}{\texttt{DKGF(W-I)}\xspace}
\newcommand{\icewsyago}{\texttt{DKGF(Y-I)}\xspace}

\newcommand{\drugbankwiki}{\texttt{DKGF(W-Drug)}\xspace}
\newcommand{\umlswiki}{\texttt{DKGF(W-UMLS)}\xspace}
\newcommand{\openalexdbpedia}{\texttt{DKGF(D-Alex)}\xspace}
\newcommand{\crunchbasedbpedia}{\texttt{DKGF(D-Crunch)}\xspace}

\newcommand{\StringMatch}{\textsf{StringMatch-F}\xspace}
\newcommand{\TFIDF}{\textsf{TF-IDF-F}\xspace}

\newcommand{\TransEF}{\textsf{TransE-F}\xspace}
\newcommand{\TransHF}{\textsf{TransH-F}\xspace}
\newcommand{\DistMultF}{\textsf{DistMult-F}\xspace}
\newcommand{\ComplExF}{\textsf{ComplEx-F}\xspace}

\newcommand{\GCNF}{\textsf{GCN-F}\xspace}

\newcommand{\TransGNNF}{\textsf{TransGNN-F}\xspace}
\newcommand{\GraphMambaF}{\textsf{Graph-Memba-F}\xspace}

\newcommand{\BERTF}{\textsf{BERT-F}\xspace}
\newcommand{\ICLF}{\textsf{ICL-F}\xspace}
\newcommand{\SelfConsistencyF}{\textsf{Self-Consistency-F}\xspace}
\newcommand{\SelfRAGF}{\textsf{Self-RAG-F}\xspace}

\newcommand{\SimpleHHEAF}{\textsf{SimpleHHEA-F}\xspace}
\newcommand{\ChatEAF}{\textsf{ChatEA-F}\xspace}

\newcommand{\KGBERTF}{\textsf{KG-BERT-F}\xspace}
\newcommand{\KGLLaMAF}{\textsf{KG-LLaMA-F}\xspace}
\newcommand{\KoPAF}{\textsf{KoPA-F}\xspace}

\newcommand{\PKGC}{\textsf{PRGC-F}\xspace}
\newcommand{\NoGenBARTF}{\textsf{NoGen-BART-F}\xspace}
\newcommand{\NoGenTfiveF}{\textsf{NoGen-T5-F}\xspace}

\newtheorem{proposition}{Proposition}

\newcommand{\squishlist}{
 \begin{list}{$\bullet$}
 { \setlength{\itemsep}{0pt}
   \setlength{\parsep}{3pt}
   \setlength{\topsep}{3pt}
   \setlength{\partopsep}{0pt}
   \setlength{\leftmargin}{1.2em}
   \setlength{\labelwidth}{1em}
   \setlength{\labelsep}{0.6em}
 }
}
\newcommand{\squishend}{
 \end{list}
}

\makeatletter

\newcommand{\Rmnum}[1]{\expandafter\@slowromancap\romannumeral #1@}
\makeatother

\makeatother

\setcopyright{acmlicensed}
\copyrightyear{xxx}
\acmYear{xxx}
\acmDOI{XXXXXXX.XXXXXXX}
\acmISBN{978-1-4503-XXXX-X/2027/05}
\acmConference[xxx]{xxx}{xxx}{xxx}
\acmBooktitle{xxx}

\begin{document}

\title{Panning for Gold: Expanding Domain-Specific Knowledge Graphs with General Knowledge}

\author{Runhao Zhao}
\affiliation{%
  \institution{National Key Laboratory of Big Data and Decision, National University of Defense Technology}
  \city{Changsha}
  \country{China}
}
\email{runhaozhao@nudt.edu.cn}

\author{Weixin Zeng}
\affiliation{%
  \institution{National Key Laboratory of Big Data and Decision, National University of Defense Technology}
  \city{Changsha}
  \country{China}
}
\email{zengweixin13@nudt.edu.cn}

\author{Wentao Zhang}
% \authornote{Corresponding authors.}
\affiliation{%
  \institution{The Center for machine learning research, Peking University}
  \city{Beijing}
  \country{China}
}
\email{wentao.zhang@pku.edu.cn}

\author{Chong Chen}
% \authornote{Corresponding authors.}
\affiliation{%
  \institution{Tsinghua University}
  \city{Beijing}
  \country{China}
}
\email{cc17@mails.tsinghua.edu.cn}

\author{Zhengpin Li}
\affiliation{%
  \institution{The Center for machine learning research, Peking University}
  \city{Beijing}
  \country{China}
}
\email{zpli@pku.edu.cn}

\author{Xiang Zhao}
\affiliation{%
  \institution{National Key Laboratory of Big Data and Decision, National University of Defense Technology}
  \city{Changsha}
  \country{China}
}
\email{xiangzhao@nudt.edu.cn}

\author{Lei Chen}
\affiliation{%
  \institution{The Hong Kong University of Science and Technology}
  \city{Clear Water Bay, Hong Kong}
  \country{China}
}
\email{leichen@cse.ust.hk}

\renewcommand{\shortauthors}{Runhao Zhao et al.}

\begin{abstract}
Domain-specific knowledge graphs (DKGs) are critical yet often suffer from limited coverage compared to General Knowledge Graphs (GKGs). Existing tasks to enrich DKGs rely primarily on extracting knowledge from external unstructured data or completing KGs through internal reasoning, but the scope and quality of such integration remain limited. This highlights a critical gap: little systematic exploration has been conducted on how comprehensive, high-quality GKGs can be effectively leveraged to supplement DKGs.
To address this gap, we propose a new and practical task: domain-specific knowledge graph fusion (DKGF), which aims to mine and integrate relevant facts from general knowledge graphs into domain-specific knowledge graphs to enhance their completeness and utility. Unlike previous research, this new task faces two key challenges: \textcircled{1} \emph{high ambiguity of domain relevance}, i.e., difficulty in determining whether knowledge from a GKG is truly relevant to the target domain, and \textcircled{2} \emph{cross-domain knowledge granularity misalignment}, i.e., GKG facts are typically abstract and coarse-grained, whereas DKGs frequently require more contextualized, fine-grained representations aligned with particular domain scenarios.
To address these, we present \ourmodel, a neuro-symbolic framework based on a novel \emph{Fact-as-Program} paradigm. \ourmodel treats fusion as an executable process, utilizing neuro-symbolic execution to infer logical relevance beyond surface similarity and employing target space grounding to calibrate granularity. We construct six new datasets to establish the first standardized evaluation suite for this task. Extensive experiments demonstrate that \ourmodel effectively overcomes domain barriers to achieve superior fusion performance.
\end{abstract}

\begin{CCSXML}
<ccs2012>
<concept>
<concept_id>10010147.10010178.10010187</concept_id>
<concept_desc>Computing methodologies~Knowledge representation and reasoning</concept_desc>
<concept_significance>500</concept_significance>
</concept>
<concept>
<concept_id>10010147.10010257.10010293.10010294</concept_id>
<concept_desc>Computing methodologies~Neural networks</concept_desc>
<concept_significance>300</concept_significance>
</concept>
<concept>
<concept_id>10002951.10002952.10003219</concept_id>
<concept_desc>Information systems~Information integration</concept_desc>
<concept_significance>500</concept_significance>
</concept>
<concept>
<concept_id>10002951.10003227.10003351</concept_id>
<concept_desc>Information systems~Data mining</concept_desc>
<concept_significance>300</concept_significance>
</concept>
</ccs2012>
\end{CCSXML}

\ccsdesc[500]{Computing methodologies~Knowledge representation and reasoning}
\ccsdesc[300]{Computing methodologies~Neural networks}
\ccsdesc[500]{Information systems~Information integration}
\ccsdesc[300]{Information systems~Data mining}

\keywords{Domain-specific Knowledge Graph Fusion; Knowledge Graph Enrichment; General-to-domain Knowledge Transfer; Fact-as-Program}

\maketitle

\section{Introduction}\label{sec:introduction}
Domain-specific knowledge graphs (DKGs) have emerged as essential infrastructures for representing structured knowledge in specialized domains and enabling intelligent applications~\cite{kdd25dkg,GeislerPVLDB25}. For example, Web-scope ICEWS\footnote{\url{https://www.andybeger.com/icews}} captures political events for international relations analysis, while PubChem\footnote{\url{https://pubchem.ncbi.nlm.nih.gov}} provides chemical knowledge widely used in drug discovery. These DKGs demonstrate the great potential of domain-specific structured knowledge in powering real-world applications (e.g., Web search and Answering)~\cite{kg22}. Nevertheless, compared with well-maintained general knowledge graphs (GKGs) such as Wikipedia~\cite{DBLP:conf/semweb/ErxlebenGKMV14} and YAGO~\cite{DBLP:conf/www/SuchanekKW07}, which benefit from community-driven contributions and timely updates, DKGs often suffer from limited completeness. They inherently face challenges due to restricted data sources~\cite{kdd25dkg}, domain-specific construction methods~\cite{KondylakisVLDB25}, and costly maintenance efforts~\cite{KhanSIGMOD25}. These limitations create a substantial barrier for DKGs to fully support downstream applications in specialized domains~\cite{KGR17}. Therefore, a pressing challenge is to efficiently and accurately enrich DKGs, improving their coverage and the ability to support domain-specific applications.

\begin{figure*}[t]
	\centering
    \includegraphics[width=1\textwidth]{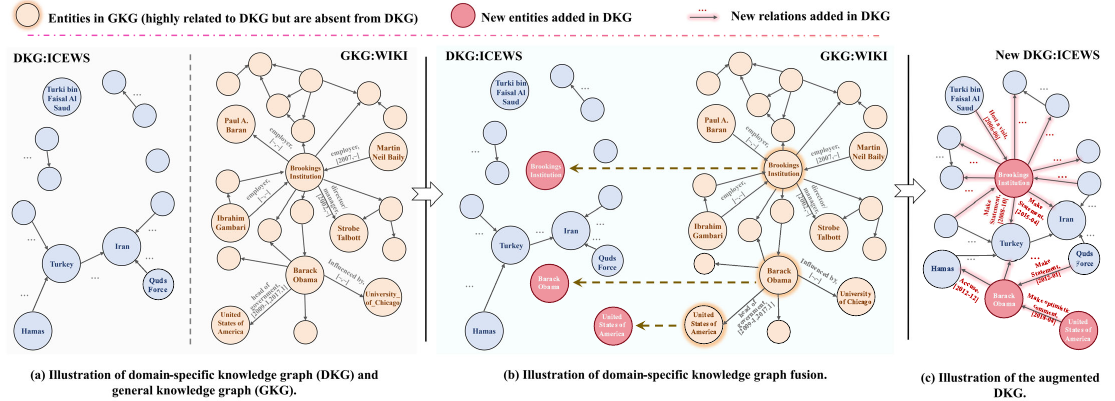}
    % \vspace{-15pt}
	\caption{Examples of domain-specific knowledge graph fusion. \autoref{fig:intro}(b) demonstrates the fusion of highly relevant entities from GKG into DKG. \autoref{fig:intro}(c) shows the augmented DKG.}
	\label{fig:intro}
    % \vspace{-10pt}
\end{figure*}

Although currently there is no work that directly investigates how to enrich DKGs, adapting existing techniques developed for supplementing GKGs may be a viable solution. Specifically, these efforts can be broadly divided into two categories: extraction from external unstructured sources and reasoning based on internal information. \textcircled{1} The former category aims to acquire new knowledge from external unstructured texts, mainly including manual annotation, entity linking, and relation triple extraction~\cite{link21,tripleSIGIR}. However, as discussed earlier, domain-specific unstructured data sources are inherently limited in scope and coverage, constraining the effectiveness of these text-based extraction methods. Moreover, the specialized terminology and complex linguistic patterns in domain-specific texts further exacerbate the challenges of accurate knowledge extraction. \textcircled{2} The latter category focuses on mining implicit knowledge from the existing graph itself, typically via knowledge graph completion~\cite{ICDEkgc24}, which predicts missing triples through internal inference within the existing KG. While these solutions can discover implicit relationships, the knowledge they generate is fundamentally bounded by the original KG's scope and cannot introduce genuinely new external knowledge. Given the limitations of this existing research and the rich, continuously updated knowledge available in comprehensive Web-scale GKGs like Wikipedia and YAGO, a natural question arises: \emph{Can we systematically leverage the abundant knowledge in GKGs to supplement and enrich DKGs?}

To this end, we propose a new and important task setting, i.e., \textbf{
\emph{domain-specific knowledge graph fusion (DKGF)}}, which aims to mine the highly relevant knowledge in the general knowledge graph to enhance the knowledge coverage of domain-specific knowledge graph, as illustrated in \autoref{fig:intro}. 
Unlike the widely studied knowledge graph alignment task~\cite{chatea}, which focuses on identifying entities that are simultaneously present and semantically equivalent across different KGs, the DKGF task essentially focuses on \emph{how to extract knowledge from a large-scale semantically complex GKG that is closely related to the target domain but have not yet been included in the DKG, and then effectively integrate them, while maintaining the semantic consistency and structural integrity of the DKG itself}. As a result, the DKGF task poses new challenges:

\textbf{Challenge~\Rmnum{1}: High Ambiguity of Domain Relevance.} Firstly, GKGs contain a large amount of knowledge with low similarity to that in DKGs, yet such information may still provide critical factual associations for enriching DKGs in reality. As illustrated in \autoref{fig:intro}, during the mining process of relevant knowledge from GKGs, the academic institution ``\emph{Brookings~Institution}'' from Washington D.C. lacks explicit connections to entities in the political crisis event DKG (e.g., ``\emph{Iran}''), resulting in low semantic similarity between entities. Nonetheless, in reality, ``\emph{Brookings~Institution}'' maintains many important connections with DKG entities and is highly likely to constitute factual associations with them. This discrepancy creates ambiguity in determining whether knowledge from GKGs is truly domain-relevant, making it difficult to directly apply traditional graph matching methods or language models~\cite{chatea} to this new task.

\textbf{Challenge~\Rmnum{2}: Cross-Domain Knowledge Granularity Misalignment.} Secondly, facts associated with relevant knowledge from GKGs often appear at levels of abstraction that differ from the granularity required by DKGs, thereby making direct fusion inappropriate and even disruptive. GKGs typically encode broad, abstract facts (e.g., ``\emph{[Barack~Obama}, \emph{head of government}, \emph{United~States}~\\\emph{of~America]}''), whereas DKGs frequently require more contextually specific, fine-grained representations aligned with particular domain tasks (e.g., ``\emph{[Barack}~\emph{Obama}, \emph{make optimistic comment}, \emph{United~States~of~America]}'' in \autoref{fig:intro}(c)). The latter can be regarded as a contextual instantiation under the broader fact of ``\emph{head of government}''. Simply transferring abstract general knowledge into a DKG can introduce semantic inconsistencies and structural misalignment. Closing this granularity gap therefore requires mechanisms to transform abstract GKG facts into domain-specific representations while preserving the semantic coherence and structural integrity of the target DKG.

In recent years, neuro-symbolic reasoning, which couples the representational power of neural networks with the explicit, verifiable semantics of symbolic logic, has attracted increasing attention for enabling inference beyond surface-level similarity and calibrating cross-domain knowledge~\cite{nstkdd24,nskg24,kdd25nskg,book24kgns}. Motivated by this capability, to address these challenges, we propose \ourmodel, a neuro-symbolic framework that reframes fusion not as static matching, but as a \textbf{\emph{Fact-as-Program}} process where knowledge facts are modeled as executable programs transformable against target domain logic. Specifically, to resolve the \emph{high ambiguity of domain relevance}, \ourmodel introduces \emph{neuro-symbolic program execution}, which maps discrete GKG facts into a continuous space and treats logic rules as transition operators to infer logical reachability beyond surface dissimilarity. Simultaneously, to address \emph{cross-domain knowledge granularity misalignment}, we employ a \emph{target space grounding} mechanism to calibrate knowledge specificity. This approach constructs a target state space and utilizes an executability core to verify that the transformed knowledge successfully grounds into the valid semantic region of the DKG, ensuring that the fused facts are not only logically relevant but also compatible with the specific granularity and structural patterns of the target domain. Notably, to fill in the gap of DKGF benchmarks, we construct six new datasets and offer 21 benchmark configurations for systematic evaluation. Extensive experiments highlight the value of the new task and demonstrate the effectiveness of \ourmodel, providing the first standardized evaluation suite for this emerging task.

\myparagraph{Contribution} We summarize our contributions as follows:

% \squishlist
% \item \underline{\emph{New Task Setting.}} 
% We define a new task setting, \emph{domain-specific knowledge graph fusion (DKGF)}, highlighting its distinct requirements compared to traditional tasks.

% \item \underline{\emph{Novel Paradigm}} We propose the \textit{Fact-as-Program} paradigm, a neuro-symbolic approach that models fact fusion as a verifiable logical execution process in a quasi-symbolic space.

% \item \underline{\emph{Theoretical Guarantee.}} We provide theoretical analysis to guarantee the expressiveness of our neuro-symbolic mapping and the validity of the granularity calibration.

% \item \underline{\emph{Comprehensive Benchmarks \& Extensive Experiments.}} We develop six new benchmark datasets to establish a standardized evaluation for the DKGF task. Comprehensive experiments demonstrate the effectiveness, efficiency, and generalization of our \ourmodel framework.
% \squishend

\squishlist
\item \underline{\emph{A New and Important Research Task.}} We define a new task setting, \emph{domain-specific knowledge graph fusion (DKGF)}, highlighting its distinct requirements compared to traditional tasks.

\item \underline{\emph{A Simple and Effective Paradigm.}} We propose the \textit{Fact-as-Program} paradigm, a neuro-symbolic approach that models fact fusion as a verifiable logical execution process in a quasi-symbolic space.

\item \underline{\emph{Theoretical Analysis.}} We provide theoretical analysis to guarantee the expressiveness of our neuro-symbolic mapping and the validity of the granularity calibration.

\item \underline{\emph{Standardized Benchmarks \& Evaluation.}} We develop six new benchmark datasets to establish a standardized evaluation for the DKGF task. Comprehensive experiments demonstrate the effectiveness, efficiency, and generalization of our \ourmodel framework.
\squishend
% \vspace{-15pt}

\section{Related work}%1小时
To the best of our knowledge, DKGF is a new task in domain-specific data integration. Closely related research can be categorized into extraction from external unstructured sources, internal knowledge reasoning, and cross-graph entity alignment.

\myparagraph{Extraction from External Unstructured Sources}
This category focuses on acquiring structured knowledge from external unstructured texts to enrich KGs. \textcircled{1} \textbf{Relation Triple Extraction (RTE)} aims to extract structured triplets from unstructured text~\cite{ragwww25}. It has evolved from supervised pipelines to zero-shot and LLM-based extraction~\cite{ragwww25,TGM25}. \textcircled{2} \textbf{Entity Linking (EL)} identifies entity mentions in text and links them to corresponding entities in KGs. While early works relied on handcrafted features~\cite{AIR25}, the field has moved towards dense retrieval and generative paradigms~\cite{link21,DeCao2021}. However, both RTE and EL heavily rely on the availability and quality of domain-specific corpora, which are often scarce or noisy in specialized domains.

\myparagraph{Internal Knowledge Reasoning}
Unlike extraction methods that seek external information, this category focuses on inferring missing facts solely based on existing information within the KG. Research in this area primarily centers on knowledge graph completion (KGC), which predicts missing triples within a single KG. Embedding-based techniques~\cite{transE,DBLP:conf/iclr/RotatE} project elements into continuous spaces. PLM-based and generative methods~\cite{KG-BERT,PKGC} further enhance performance. Although effective for link prediction, KGC is inherently limited to the closed-world assumption of the existing graph and cannot discover new entities or facts from the outside world.

\myparagraph{Entity Alignment} 
Entity alignment (EA) aims to identify equivalent entities across KGs. Early translation-based methods~\cite{MTransE} utilize distance models but often struggle with complex structures. GNN-based approaches~\cite{GCN-Align,Dual-AMN,TEA-GNN,STEA,dualmatch} leverage graph convolutions to capture neighborhood topologies, while recent solutions integrate side information~\cite{simplehhea} or leverage LLMs for few-shot semantic matching~\cite{llm4ea,chatea,AdaCoAgent}.

\myparagraph{Task Distinction and Gap Analysis} Despite these efforts, existing research predominantly relies on extracting knowledge from unstructured data or completing KGs through internal reasoning, while the scope and quality of knowledge integration remain significantly limited~\cite{DuetGraph25}. This situation exposes a critical gap: There has been little systematic exploration of how comprehensive, high-quality GKGs can be effectively leveraged to supplement DKGs. DKGF is therefore an essential and indispensable task. Furthermore, this new task differs fundamentally from existing ones: \textcircled{1} \textbf{Vs. RTE.} RTE operates on \emph{unstructured text}, whereas DKGF leverages \emph{structured GKG}. \textcircled{2} \textbf{Vs. EL.} EL resolves \emph{text-to-KG disambiguation}, whereas DKGF enhances \emph{DKG coverage via cross structured graph discovery and fusion}. \textcircled{3} \textbf{Vs. KGC.} KGC focuses on reasoning within \emph{a single KG}, while DKGF integrates knowledge from \emph{external GKG}. \textcircled{4} \textbf{Vs. EA.} EA outputs \emph{same entity pairs} across KGs, whereas DKGF identifies \emph{missing but relevant entities and produces new facts}. Consequently, existing research cannot address the core challenges of DKGF: \emph{high ambiguity of domain relevance} and \emph{cross-domain granularity misalignment}. While neuro-symbolic reasoning~\cite{nstkdd24,nskg24,kdd25nskg,book24kgns} provides the explicit logical grounding needed to resolve such ambiguity, it has typically been confined to single-graph inference. Inspired by this, our \ourmodel adopts a \emph{Fact-as-Program} paradigm, extending neuro-symbolic execution to dynamically validate logical relevance and granularity beyond surface matching.

\section{Problem Statement}\label{sec:problem}

\myparagraph{Domain-specific Knowledge Graph}
The domain-specific knowledge graphs $G^d = (V^d, R^d, F^d)$ and general knowledge graphs $G^g = (V^g, R^g, F^g)$ encapsulate real-world domain-specific and general knowledge through three core components: entity sets $V^d, V^g$, relation sets $R^d, R^g$, and fact sets $F^d, F^g$~\cite{kg22}. Each fact in $F^d$ and $F^g$ is represented in the form of $(v^d, r^d, \hat{v}^d) \in V^d \times R^d \times V^d$ and $(v^g, r^g, \hat{v}^g) \in V^g \times R^g \times V^g$, respectively, where entities $v^d, \hat{v}^d \in V^d$, $v^g, \hat{v}^g \in V^g$, relations $r^d \in R^d$, and $r^g \in R^g$.

\myparagraph{Domain-specific Knowledge Graph Fusion}
Based on the aforementioned definitions, the domain-specific knowledge graph fusion task aims to mine the highly relevant knowledge in the general knowledge graph $G^g = \big(V^g, R^g, F^g\big)$ to enhance the knowledge coverage of domain-specific knowledge graph $G^d=\big(V^d, R^d, F^d\big)$. Formally, given a DKG $G^d$ and GKG $G^g$, the DKGF task generates a new set of fused facts $F_{fused}$ for $G^d$: 
\begin{equation}
% \resizebox{0.9\hsize}{!}{
F_{fused} = \{(v, r, \hat{v}) \mid\; v, \hat{v} \in V^d \cup V^g,\; r \in R^d, (v, r, \hat{v}) \notin F^d \},
% }
\label{fused}
\end{equation}
where $(v, r, \hat{v})$ represents a triplet with head entity $v$, relation $r$, and tail entity $\hat{v}$.

\begin{figure*}[t]
	\centering
    \includegraphics[width=0.85\textwidth]{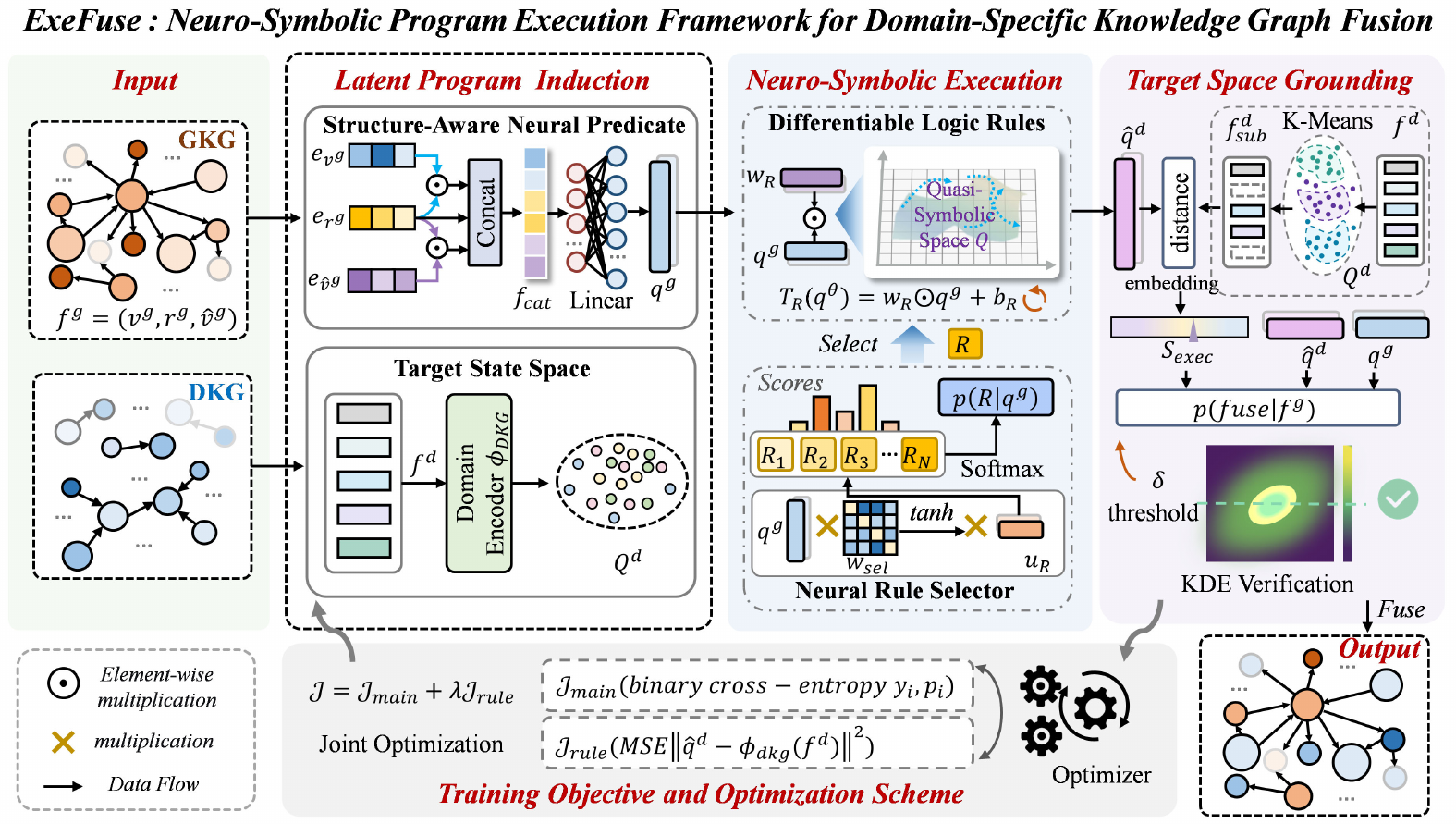}
    % \vspace{-10pt}
	\caption{The framework of \ourmodel for DKGF.}
	\label{fig:method1}
    % \vspace{-13pt}
\end{figure*}

\section{Methodology}
\label{sec:method}
In this section, we present \ourmodel, a neuro-symbolic framework designed to address the unique challenges of \emph{domain-specific knowledge graph fusion (DKGF)}.
As analyzed in Section~\ref{sec:introduction}, the core difficulty of DKGF lies in \emph{high ambiguity of domain relevance} and \emph{cross-domain granularity misalignment}.
Existing static similarity matching fails to bridge this gap. Instead, effective fusion requires answering two fundamental questions:
\emph{(i) how to uncover implicit logical connections hidden beneath surface-level dissimilarity}, and
\emph{(ii) how to adapt abstract general facts to the specific granularity of the target domain}. To this end, we propose the \textbf{\emph{Fact-as-Program}} paradigm, which reframes fusion not as a classification task, but as a \textbf{dynamic execution process}~\cite{bs25,nstkdd24,kdd25nskg,book24kgns}.

As illustrated in \autoref{fig:method1}, \ourmodel consists of three coordinated stages. \textcircled{1} \textbf{Stage 1: Latent Program Induction.} To handle structural heterogeneity, we first encode discrete GKG facts into continuous program states using a structure-aware predicate, preparing them for logical transformation. \textcircled{2} \textbf{Stage 2: Resolving Ambiguity via Neuro-Symbolic Execution.} To resolve the ambiguity of relevance, we move beyond surface embedding similarity. We treat logic rules as transition operators that ``execute'' the fact. By applying these operators, we infer the \emph{logical reachability} of a fact—transforming it from its original state to a potential domain-relevant state. This ensures that even semantically distant entities can be linked if there exists a valid logical path. \textcircled{3} \textbf{Stage 3: Granularity Calibration via Target Space Grounding.} To address granularity misalignment, we view the execution result as a proposal that must be verified. We construct a \emph{Target State Space} representing the valid semantic distribution of the DKG. The ``program'' is considered successfully executed only if its output can be grounded (i.e., mapped with high probability) into this specific target region, ensuring the fused knowledge aligns with the domain's granularity.

\subsection{Formalization \& Overview}
\label{sec:overview}
In this paradigm, we treat each source GKG fact $f^g = (v^g, r^g, \hat{v}^g)$ as a latent executable program $\Pi(f^g)$ defined in a continuous \textbf{Quasi-Symbolic Space} $\mathcal{Q} \subseteq \mathbb{R}^d$. Formally, the framework is a tuple $\Pi = (\phi_{\text{enc}}, \mathcal{L}, \mathcal{Q})$, where \textcircled{1} $\phi_{\text{enc}}$ is the \textbf{Structure-Aware Neural Predicate} that maps discrete GKG facts into continuous program states, preserving structural semantics; \textcircled{2} $\mathcal{L}$ is the set of \textbf{Inductive Logic Rules} acting as transition operators to transform states into domain states; and \textcircled{3} $\mathcal{Q}$ is the differentiable state space where geometric transformations correspond to logical entailments. A GKG fact $f^g$ is deemed valid for fusion into the DKG $G^d$ if its program is \textit{executable}. Executability is defined as the ability to transition from an initial state $\mathbf{q}^g$ to a valid target state $\mathbf{q}^d$ that resides within the \textbf{Target State Space} of the DKG.

\subsection{Latent Program Induction via Structure-Aware Predicates}
\label{sec:encoding}

The first stage compiles the discrete symbolic fact into an initial executable state. Unlike standard approaches that treat facts as flat feature vectors, we propose a \textbf{Structure-Aware Neural Predicate} to capture the compositional logic of the triplet $(v^g, r^g, \hat{v}^g)$.

\subsubsection{Structure-Aware Neural Predicate} Let $\mathbf{E}^g \in \mathbb{R}^{|V^g| \times d_{emb}}$ and $\mathbf{R}^g \in \mathbb{R}^{|R^g| \times d_{emb}}$ be the pre-trained embedding matrices. We define the initial program state $\mathbf{q}^g \in \mathcal{Q}$ via a relational composition function:
\begin{equation}
    \label{eq:encoding}
    \resizebox{0.9\hsize}{!}{$
    \mathbf{q}^g = \phi_{\text{enc}}(f^g) = \mathbf{W}_2 \cdot \text{GeLU}\left( \mathbf{W}_1 \left[ (\mathbf{e}_{v^g} \odot \mathbf{e}_{r^g}) \,;\, (\mathbf{e}_{\hat{v}^g} \odot \mathbf{e}_{r^g}) \,;\, \mathbf{e}_{r^g} \right] + \mathbf{b}_1 \right),$}
\end{equation}
where $\mathbf{e}_{(\cdot)}$ denotes the embedding vector retrieved from $\mathbf{E}^g$ or $\mathbf{R}^g$, $\odot$ represents the Hadamard (element-wise) product, and $[;]$ denotes concatenation. $\mathbf{W}_1, \mathbf{W}_2$ and $\mathbf{b}_1$ are learnable parameters.
\textbf{Remark:} The interaction terms $(\mathbf{e}_{v} \odot \mathbf{e}_{r})$ and $(\mathbf{e}_{\hat{v}} \odot \mathbf{e}_{r})$ explicitly model the interaction between the relation and the head/tail entities, ensuring that $\mathbf{q}^g$ encodes the \textit{structural validity} of the fact.

\subsubsection{Target State Space} We define the \textbf{Target State Space} $\mathcal{Q}^d \subset \mathcal{Q}$ as the semantic region populated by the embeddings of all existing valid DKG facts:
\begin{equation}
    \mathcal{Q}^d = \{ \phi_{\text{dkg}}(f^d) \mid f^d \in F^d \},
\end{equation}
where $\phi_{\text{dkg}}$ is the domain-specific encoder. It shares the same architecture as $\phi_{\text{enc}}$ but is fine-tuned to represent the target domain distribution.

\subsection{Stage 2: Resolving Ambiguity via Neuro-Symbolic Execution}
\label{sec:execution}

This stage addresses \textbf{Challenge \Rmnum{1} (Ambiguity of Domain Relevance)}.
In a static space, relevance is often ambiguous because direct similarity fails to capture implicit connections. We argue that relevance is best established through \textbf{logical reachability}~\cite{nskg24}: if a fact implies a valid domain state through explicit reasoning rules, the ambiguity is resolved.

\subsubsection{Differentiable Logic Rules}
We learn a set of logic rules $\mathcal{L}$. Symbolically, a rule $R \in \mathcal{L}$ implies $r^d(X,Y) \leftarrow r^g(X,Y) \wedge \mathcal{C}$.
To support differentiable execution, we parameterize each rule $R$ as an affine transformation. The rule application function $\mathcal{T}_R: \mathcal{Q} \to \mathcal{Q}$ is defined as: $\mathcal{T}_R(\mathbf{q}^g) = \mathbf{w}_R \odot \mathbf{q}^g + \mathbf{b}_R$,
where $\mathbf{w}_R, \mathbf{b}_R \in \mathbb{R}^d$ are learnable parameters.
Geometrically, this transformation models the \emph{inference step}, where $\mathbf{w}_R$ and $\mathbf{b}_R$ adjust the semantic focus to trace the logical path from the source fact to the target context.

\subsubsection{Neural Rule Selector}
To address the ambiguity of which rule applies, we employ a Neural Rule Selector that computes an attention distribution $p(R \mid \mathbf{q}^g)$ over $\mathcal{L}$:
\begin{equation}
    \label{eq:selector}
    p(R \mid \mathbf{q}^g) = \frac{\exp(\mathbf{u}_R^\top \tanh(\mathbf{W}_{sel} \mathbf{q}^g))}{\sum_{R' \in \mathcal{L}} \exp(\mathbf{u}_{R'}^\top \tanh(\mathbf{W}_{sel} \mathbf{q}^g))},
\end{equation}
where $\mathbf{W}_{sel} \in \mathbb{R}^{d \times d}$ projects the state into the rule-selection space, and $\mathbf{u}_R \in \mathbb{R}^d$ is a rule-specific query vector.

\subsubsection{Program Execution}
The executed program state $\hat{\mathbf{q}}^d$ is the expected outcome of applying the rules:
\begin{equation}
    \label{eq:execution}
    \hat{\mathbf{q}}^d = \sum_{R \in \mathcal{L}} p(R \mid \mathbf{q}^g) \cdot \mathcal{T}_R(\mathbf{q}^g).
\end{equation}

By explicitly executing these logic rules, \ourmodel bridges the ambiguity gap via logical inference, recovering implicit connections that pure similarity search would miss.

\subsection{Stage 3: Granularity Calibration via Target Space Grounding}
\label{sec:decoding}

This stage addresses \textbf{Challenge \Rmnum{2} (Granularity Misalignment)}.
While logic execution ensures relevance, the resulting state $\hat{\mathbf{q}}^d$ may still be too abstract (e.g., generic concepts) to fit the DKG. To solve this, we perform \textbf{Granularity Calibration} by checking if the executed state $\hat{\mathbf{q}}^d$ can be successfully grounded into the \textbf{Target State Space} $\mathcal{Q}^d$.

\subsubsection{Executability Score}
We define the executability score $S_{exec}(f^g)$ using kernel density estimation (KDE) over a set of DKG prototypes $F^d_{sub} \subset F^d$:
\begin{equation}
    \label{eq:exec_score}
    S_{exec}(f^g) = \frac{1}{|F^d_{sub}|} \sum_{f^d_j \in F^d_{sub}} \exp\left( - \frac{\| \hat{\mathbf{q}}^d - \phi_{\text{dkg}}(f^d_j) \|^2}{\tau} \right),
\end{equation}
where $\tau$ is a temperature hyperparameter. To ensure efficiency, $F^d_{sub}$ consists of $K$ cluster centroids derived from K-Means clustering on the DKG fact embeddings ($K \ll |F^d|$).
Intuitively, $S_{exec}$ measures whether the transformed fact ``lands'' in a semantically valid region of the Target State Space.

\subsubsection{Fusion Decision}
The final fusion probability is computed by a fusion network $\Psi$ that integrates the structural context and the verification score:
\begin{equation}
    p(\text{fuse} \mid f^g) = \sigma \left( \Psi([\mathbf{q}^g \,;\, \hat{\mathbf{q}}^d \,;\, \tilde{S}_{exec}]) \right),
\end{equation}
where $\tilde{S}_{exec}$ is the normalized executability score. Facts with $p > \delta$ are added to the final fused set $F_{fused}$. The complexity, scalability, and theoretical analysis can be found in Appendix~\ref{sec:complexity}-\ref{sec:theory}

\subsection{Training Objective}
\label{sec:optimization}

We optimize the framework using a joint loss function on a dataset of aligned pairs $\mathcal{D}_{pos}$ (constructed via distant supervision) and negative samples.

\textbf{Main Fusion Loss.} We minimize the binary cross-entropy loss $\mathcal{J}_{main}$ for the fusion decision $p(\text{fuse})$:
\begin{equation}
    \mathcal{J}_{main} = -\frac{1}{N} \sum_{i=1}^N \left[ y_i \log p_i + (1-y_i) \log (1 - p_i) \right].
\end{equation}

\textbf{Rule Consistency Loss.} To enforce valid semantic transformations, we minimize the regression loss on positive pairs: $\mathcal{J}_{rule} = \sum_{(f^g, f^d) \in \mathcal{D}_{pos}} \| \hat{\mathbf{q}}^d(f^g) - \phi_{\text{dkg}}(f^d) \|^2$. The final objective is $\mathcal{J} = \mathcal{J}_{main} + \lambda \mathcal{J}_{rule}$.

\section{Experiments}\label{sec:experiment}% 措辞未改
In this section, we introduce the experimental setup and then extensively discuss the results, highlighting the strengths of our proposed datasets and framework from multiple perspectives. The following research questions (RQ) are the focus of our experiments.
\begin{itemize} % [leftmargin=16pt]
    \item[\textbf{RQ1.}] Can \ourmodel effectively address the above challenges and achieve significant improvements? (Section \ref{exp:main})
    \item[\textbf{RQ2.}] Does the new task offer sufficient research value, and do the benchmark datasets provide meaningful insights for the solution to this new task?  (Section \ref{exp:main})
    \item[\textbf{RQ3.}] How much do each module in the {\ourmodel} contribute to the final results? Are these modules reasonably designed? (Section \ref{exp:abl})
    \item[\textbf{RQ4.}] Does the \ourmodel successfully balance alignment accuracy and efficiency? (Section \ref{exp:efficiency})
\end{itemize}

\begin{table}[t!]
    \centering
    % 调高行高倍数
    \caption{Dataset statistics across diverse domains. ``\textit{\#$F_{fused}$}'': The total number of fused new facts (tuples) from GKG into DKG. ``\textit{Structure. Sim.}'': The average neighbor structure similarity of entities in DKG and GKG, as defined in \cite{simplehhea}.}
    \label{tab:stats}
    \begin{adjustbox}{max width=0.9\columnwidth}
    \begin{tabular}{l| l l c c}
        \toprule
        \bfseries Domain & \bfseries Dataset & \bfseries Graph Source & \bfseries \#$F_{fused}$ & \bfseries Structure. Sim. \downgood \\
        \midrule
        \multirow{4}{*}{\textbf{Political}} & \multirow{2}{*}{\icewswiki} & DKG: ICEWS & \multirow{2}{*}{796,254} & \multirow{2}{*}{15.4\%} \\
        & & GKG: Wikidata & & \\
        \cmidrule{2-5}
        & \multirow{2}{*}{\icewsyago} & DKG: ICEWS & \multirow{2}{*}{451,158} & \multirow{2}{*}{14.0\%} \\
        & & GKG: YAGO & & \\
        \midrule
        \multirow{4}{*}{\textbf{Biomedical}} & \multirow{2}{*}{\drugbankwiki} & DKG: DrugBank & \multirow{2}{*}{532,810} & \multirow{2}{*}{18.2\%} \\
        & & GKG: Wikidata & & \\
        \cmidrule{2-5}
        & \multirow{2}{*}{\umlswiki} & DKG: UMLS & \multirow{2}{*}{1,204,561} & \multirow{2}{*}{12.5\%} \\
        & & GKG: Wikidata & & \\
        \midrule
        \multirow{2}{*}{\textbf{Academic}} & \multirow{2}{*}{\openalexdbpedia} & DKG: OpenAlex & \multirow{2}{*}{2,105,340} & \multirow{2}{*}{9.8\%} \\
        & & GKG: DBpedia & & \\
        \midrule
        \multirow{2}{*}{\textbf{Business}} & \multirow{2}{*}{\crunchbasedbpedia} & DKG: Crunchbase & \multirow{2}{*}{642,109} & \multirow{2}{*}{11.3\%} \\
        & & GKG: DBpedia & & \\
        \bottomrule
    \end{tabular}
    \end{adjustbox}
\end{table}

\begin{table*}[t!]
	\caption{Main experiment results on \icewswiki and \icewsyago datasets (\textbf{General-purpose configurations}). We report the accuracy (ACC), precision (P), recall (R), and F1-score (F1). The best overall results are highlighted in \textbf{bold}, while the best baseline results within this category are \underline{underlined}. ``\textit{Sema.}, \textit{Struc.}, \textit{LLM.}'' indicate the use of semantics, structural information, and large language models, respectively. Relative performance improvements over the best baseline are indicated in \textbf{\textcolor{green!70!black}{green} \upgood}. These conventions apply to all subsequent tables.}
	\label{tb:main_results_general}
	\centering
	\begin{adjustbox}{max width=0.95\textwidth}
		\begin{tabular}{c|l|p{0.5cm}p{0.5cm}p{0.5cm}|>{\columncolor{lightgray!50}}c|cc|>{\columncolor{lightgray!50}}c||>{\columncolor{lightgray!50}}c|cc|>{\columncolor{lightgray!50}}c||>{\columncolor{lightgray!50}}c|cc|>{\columncolor{lightgray!50}}c||>{\columncolor{lightgray!50}}c|cc|>{\columncolor{lightgray!50}}c}
			\toprule
			\multicolumn{2}{c|}{\multirow{2}{*}{\textbf{Benchmark Configurations}}} &\multicolumn{3}{c|}{\textbf{Settings}}&\multicolumn{4}{c||}{\textbf{\icewswiki-S1}} &\multicolumn{4}{c||}{\textbf{\icewsyago-S1}} &\multicolumn{4}{c||}{\textbf{\icewswiki-S2}} &\multicolumn{4}{c}{\textbf{\icewsyago-S2}}\cr
			\cmidrule(lr){3-5}\cmidrule(lr){6-9}\cmidrule(lr){10-13}\cmidrule(lr){14-17}\cmidrule(lr){18-21}
			\multicolumn{2}{c|}{}&\textit{Sema.}&\textit{Struc.}&\textit{LLM.}&ACC &P &R &F1 &ACC &P &R &F1&ACC &P &R &F1&ACC &P &R &F1\cr
			\midrule

			\multirow{2}{*}{\rotatebox{90}{Rule.}} &\StringMatch&\checkmark&&& 0.496  & 0.160  & 0.002  & 0.004  & 0.498  & 0.286  & 0.003  & 0.005  & 0.498  & 0.192  & 0.002  & 0.003  & 0.498  & 0.308  & 0.003  & 0.005\cr
			&\TFIDF&\checkmark&&& 0.498  & 0.461  & 0.025  & 0.047  & 0.507  & 0.598  & 0.043  & 0.081  & 0.498  & 0.451  & 0.023  & 0.043  & 0.508  & 0.598  & 0.046  & 0.086\cr
            
			\midrule
            
			\multirow{4}{*}{\rotatebox{90}{Trans.}}&\TransEF&\checkmark&\checkmark&& \underline{0.649}  & 0.668  & 0.596  & \underline{0.630}  & 0.588  & 0.782  & 0.244  & 0.372  & \underline{0.637}  & 0.662  & 0.560  & \underline{0.607}  & 0.575  & 0.772  & 0.212  & 0.332\cr
			&\TransHF&\checkmark&\checkmark&& 0.640  & 0.683  & 0.520  & 0.591  & 0.583  & 0.764  & 0.240  & 0.365  & 0.632  & 0.685  & 0.489  & 0.570  & 0.568  & 0.768  & 0.195  & 0.311\cr
			&\DistMultF&\checkmark&\checkmark&& 0.554  & 0.547  & 0.629  & 0.585  & 0.525  & 0.524  & 0.548  & 0.536  & 0.536  & 0.532  & 0.590  & 0.560  & 0.532  & 0.530  & 0.562  & 0.545\cr
			&\ComplExF&\checkmark&\checkmark&& 0.553  & 0.546  & 0.632  & 0.586  & 0.526  & 0.525  & 0.567  & 0.545  & 0.544  & 0.539  & 0.609  & 0.572  & 0.515  & 0.514  & 0.546  & 0.530 \cr
   		
			\midrule
            
			\multirow{3}{*}{\rotatebox{90}{GNN.}}&\GCNF&&\checkmark&& 0.485  & 0.487  & 0.572  & 0.526  & 0.501  & 0.501  & 0.793  & 0.614  & 0.493  & 0.494  & 0.582  & 0.534  & 0.498  & 0.499  & 0.796  & 0.613\cr
			&\TransGNNF&\checkmark&\checkmark&& 0.490  & 0.491  & 0.505  & 0.498  & 0.465  & 0.465  & 0.477  & 0.471  & 0.486  & 0.487  & 0.507  & 0.497  & 0.468  & 0.468  & 0.481  & 0.474\cr
            &\GraphMambaF&\checkmark&\checkmark&& 0.485  & 0.487  & 0.572  & 0.526  & 0.503  & 0.502  & 0.794  & \underline{0.615}  & 0.493  & 0.494  & 0.582  & 0.534  & 0.499  & 0.501  & 0.796  & \underline{0.615}\cr
        
            \midrule
            
			\multirow{4}{*}{\rotatebox{90}{Gen.}}&\BERTF&\checkmark&&& 0.532  & 0.531  & 0.549  & 0.540  & 0.586  & 0.599  & 0.523  & 0.558  & 0.530  & 0.531  & 0.507  & 0.519  & 0.569  & 0.561  & 0.631  & 0.594\cr
			&\ICLF&\checkmark&&\checkmark& 0.550  & 0.553  & 0.528  & 0.540  & 0.583  & 0.573  & 0.649  & 0.609  & 0.494  & 0.494  & 0.523  & 0.508  & 0.488  & 0.489  & 0.523  & 0.505\cr
			&\SelfConsistencyF&\checkmark&\checkmark&\checkmark& 0.592  & 0.607  & 0.522  & 0.561  & 0.590  & 0.584  & 0.621  & 0.602  & 0.531  & 0.533  & 0.507  & 0.520  & 0.566  & 0.559  & 0.628  & 0.591\cr
            &\SelfRAGF&\checkmark&\checkmark&\checkmark& 0.576  & 0.589  & 0.506  & 0.544  & \underline{0.594}  & 0.623  & 0.477  & 0.540  & 0.544  & 0.544  & 0.545  & 0.545  & \underline{0.577}  & 0.595  & 0.481  & 0.532\cr
        
			\midrule
            \midrule

			\multicolumn{2}{c|}{\textbf{\ourmodel (Ours)}} &\checkmark&\checkmark&& \textbf{0.680} & 0.673 & 0.708 & \textbf{0.690}  & \textbf{0.661} & 0.615 & 0.717 & \textbf{0.662}  & \textbf{0.655} & 0.627 & 0.683 & \textbf{0.654}  & \textbf{0.633} & 0.634 & 0.682 & \textbf{0.657}\cr
            \midrule
            \multicolumn{5}{c|}{\textit{Relative Improvement (\%)}} & \textbf{\textcolor{green!70!black}{4.78\% \upgood}} & - & - & \textbf{\textcolor{green!70!black}{9.52\% \upgood}} & \textbf{\textcolor{green!70!black}{11.28\% \upgood}} & - & - & \textbf{\textcolor{green!70!black}{7.64\% \upgood}} & \textbf{\textcolor{green!70!black}{2.83\% \upgood}} & - & - & \textbf{\textcolor{green!70!black}{7.74\% \upgood}} & \textbf{\textcolor{green!70!black}{9.71\% \upgood}} & - & - & \textbf{\textcolor{green!70!black}{6.83\% \upgood}} \cr
			\bottomrule
	\end{tabular}
	\end{adjustbox}
\end{table*}

\subsection{Experimental Settings}
\subsubsection{DKGF Benchmarks and Benchmark Suite}
Currently, there is no benchmark that effectively studies and evaluates the DKGF task. To this end, we construct six new DKGF benchmark datasets across a diverse set of critical fields (Political, Biomedical, Academic, and Business) to support comprehensive knowledge integration and cross-domain evaluation (c.f.~\autoref{tab:stats}). The DKGs are ICEWS, DrugBank, UMLS, OpenAlex, and Crunchbase, while the GKGs are Wikidata, YAGO, and DBpedia. 
Since DKGF has no pre-existing specialized solutions, we establish a standardized suite of 21 configurations, grouped into (i) general-purpose configurations (Rule/Trans./GNN/Gen.) and (ii) cross-task adaptation configurations (EA/KGC/RTE). 
All methods are adapted under a unified Fusion Scoring Protocol. 
The full dataset construction details, complete baseline list, and adaptation protocol are provided in Appendix~\ref{app:Experiments}.

\begin{table*}[t!]
	\caption{Main experiment results on \icewswiki and \icewsyago datasets (\textbf{Cross-task adaptation configurations}). }
	\label{tb:main_results_crosstask}
	\centering
	\begin{adjustbox}{max width=0.95\textwidth}
		\begin{tabular}{c|l|p{0.5cm}p{0.5cm}p{0.5cm}|>{\columncolor{lightgray!50}}c|cc|>{\columncolor{lightgray!50}}c||>{\columncolor{lightgray!50}}c|cc|>{\columncolor{lightgray!50}}c||>{\columncolor{lightgray!50}}c|cc|>{\columncolor{lightgray!50}}c||>{\columncolor{lightgray!50}}c|cc|>{\columncolor{lightgray!50}}c}
			\toprule
			\multicolumn{2}{c|}{\multirow{2}{*}{\textbf{Benchmark Configurations}}} &\multicolumn{3}{c|}{\textbf{Settings}}&\multicolumn{4}{c||}{\textbf{\icewswiki-S1}} &\multicolumn{4}{c||}{\textbf{\icewsyago-S1}} &\multicolumn{4}{c||}{\textbf{\icewswiki-S2}} &\multicolumn{4}{c}{\textbf{\icewsyago-S2}}\cr
			\cmidrule(lr){3-5}\cmidrule(lr){6-9}\cmidrule(lr){10-13}\cmidrule(lr){14-17}\cmidrule(lr){18-21}
			\multicolumn{2}{c|}{}&\textit{Sema.}&\textit{Struc.}&\textit{LLM.}&ACC &P &R &F1 &ACC &P &R &F1&ACC &P &R &F1&ACC &P &R &F1\cr
			\midrule
            
            \multirow{2}{*}{\rotatebox{90}{EA.}} &\SimpleHHEAF&\checkmark&\checkmark&& 0.490 & 0.492 & 0.507 & 0.499  & 0.493 & 0.493 & 0.503 & 0.498  & 0.492 & 0.492 & 0.507 & 0.500  & 0.477 & 0.477 & 0.481 & 0.479\cr
			&\ChatEAF&\checkmark&\checkmark&\checkmark& \underline{0.596} & 0.551 & 0.581 & \underline{0.566}  & \underline{0.649}  & 0.727  & 0.477  & 0.576  & \underline{0.592} & 0.611 & 0.507 & 0.554 & \underline{0.592} & 0.619 & 0.481 & \underline{0.541}\cr
        
			\midrule
			\multirow{4}{*}{\rotatebox{90}{KGC.}}         
            &\KGBERTF&\checkmark&\checkmark&& 0.523 & 0.522 & 0.546 & 0.534  & 0.516 & 0.515 & 0.531 & 0.523  & 0.507 & 0.507 & 0.515 & 0.511 & 0.503 & 0.503 & 0.481 & 0.492\cr
			&\KGLLaMAF&\checkmark&\checkmark&\checkmark& 0.512 & 0.512 & 0.531 & 0.521  & 0.539 & 0.537 & 0.558 & 0.547  & 0.529 & 0.527 & 0.529 & 0.528 & 0.521 & 0.521 & 0.514 & 0.517\cr
			&\KoPAF&\checkmark&\checkmark&\checkmark& 0.558 & 0.556 & 0.577 & \underline{0.566}  & 0.561 & 0.556 & 0.602 & \underline{0.578}  & 0.549 & 0.547 & 0.564 & \underline{0.556} & 0.541 & 0.543 & 0.522 & 0.532\cr
            &\PKGC&\checkmark&\checkmark&& 0.503 & 0.503 & 0.508 & 0.505  & 0.507 & 0.507 & 0.515 & 0.511  & 0.504 & 0.504 & 0.507 & 0.506 & 0.503 & 0.503 & 0.481 & 0.492\cr
            
			\midrule
			\multirow{2}{*}{\rotatebox{90}{RTE.}} 
            &\NoGenBARTF&\checkmark&\checkmark&& 0.516 & 0.515 & 0.531 & 0.523  & 0.511 & 0.511 & 0.523 & 0.517  & 0.508 & 0.508 & 0.517 & 0.512 & 0.509 & 0.509 & 0.491 & 0.500\cr
			&\NoGenTfiveF&\checkmark&\checkmark&& 0.510 & 0.510 & 0.522 & 0.516  & 0.507 & 0.507 & 0.515 & 0.511  & 0.509 & 0.509 & 0.519 & 0.514 & 0.504 & 0.504 & 0.509 & 0.506\cr

			\midrule
            \midrule
            
			\multicolumn{2}{c|}{\textbf{\ourmodel (Ours)}} &\checkmark&\checkmark&& \textbf{0.680} & 0.673 & 0.708 & \textbf{0.690}  & \textbf{0.661} & 0.615 & 0.717 & \textbf{0.662}  & \textbf{0.655} & 0.627 & 0.683 & \textbf{0.654}  & \textbf{0.633} & 0.634 & 0.682 & \textbf{0.657}\cr
            \midrule
            \multicolumn{5}{c|}{\textit{Relative Improvement (\%)}} & \textbf{\textcolor{green!70!black}{14.09\% \upgood}} & - & - & \textbf{\textcolor{green!70!black}{21.91\% \upgood}} & \textbf{\textcolor{green!70!black}{1.85\% \upgood}} & - & - & \textbf{\textcolor{green!70!black}{14.53\% \upgood}} & \textbf{\textcolor{green!70!black}{10.64\% \upgood}} & - & - & \textbf{\textcolor{green!70!black}{17.63\% \upgood}} & \textbf{\textcolor{green!70!black}{6.93\% \upgood}} & - & - & \textbf{\textcolor{green!70!black}{21.44\% \upgood}} \cr
			\bottomrule
	\end{tabular}
	\end{adjustbox}
\end{table*}

\begin{figure*}[t]
	\centering
    \includegraphics[width=0.85\textwidth]{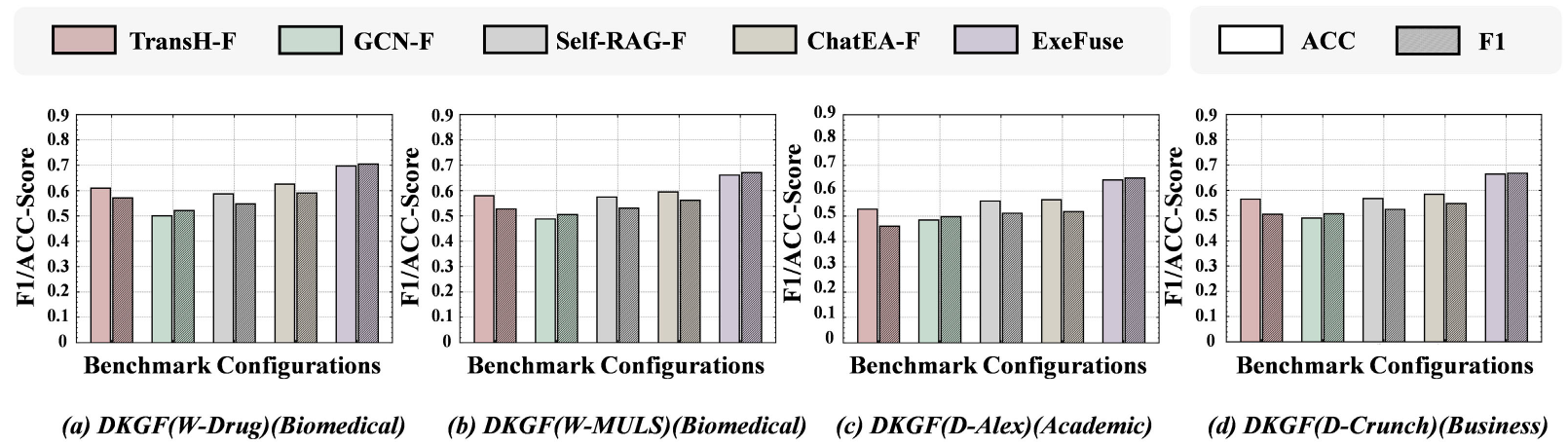}
    \vspace{-10pt}
	\caption{Comprehensive evaluation across diverse domain-specific knowledge graphs on \drugbankwiki, \umlswiki, \openalexdbpedia and \crunchbasedbpedia.}
	\label{fig:main_multi}
    \vspace{-10pt}
\end{figure*}

\subsubsection{Evaluation Protocol and Implementation Details}  To evaluate the quality of DKGF produced by \ourmodel and the benchmark configurations, we follow the triple classification task~\cite{tripleSIGIR, kgcmm24} and use accuracy (ACC), precision (P), recall (R), and F1-score (F1) as evaluation metrics. This task is essentially a binary classification problem, and all test datasets are label-balanced. We also follow two classic data split settings~\cite{kgcmm24, chatea}, where \icewswiki-S1 and \icewsyago-S1 use 80\% of training data, and \icewswiki-S2 and \icewsyago-S2 use 70\% to test robustness. Other benchmark datasets follow the S1 split by default. All methods are evaluated using a unified fusion scoring protocol.

We use runtime (in seconds) as an efficiency metric. All results are averaged over five independent runs with different random seeds. All LLMs are implemented using the same model version, gpt-4o. Additional details on design alternatives, distant supervision heuristics, baseline adaptation protocols, and parameter sensitivity analyses are provided in our anonymous repository\footnote{\url{https://anonymous.4open.science/r/DKGF_1-D85B/}\label{repository}}.

%这里到时候要详细写,因为是创新；还没写完

\subsection{Main Results (RQ1 \& RQ2)}\label{exp:main}
\subsubsection{Comparison with All Configurations (RQ1)} 
Table~\ref{tb:main_results_general} and Table~\ref{tb:main_results_crosstask} compare \ourmodel with general-purpose and cross-task adaptation baselines on DKGF, and \ourmodel achieves the best overall performance across all settings. The key observations are as follows. (1) \emph{Existing paradigms are insufficient.} General-purpose methods based on rules, translation, GNNs, or generative models, as well as methods adapted from entity alignment, knowledge graph completion, and relation triple extraction, consistently underperform \ourmodel. This indicates that DKGF cannot be solved by directly modeling semantic similarity or structural proximity, nor by simply transferring models from related KG tasks; domain-dependent execution constraints and task-specific relevance modeling are necessary. (2) \emph{Generalization across specialized domains.} On four additional datasets covering biomedical (\drugbankwiki, \umlswiki), academic (\openalexdbpedia), and business (\crunchbasedbpedia) domains, \ourmodel consistently outperforms the strongest baselines, as shown in \autoref{tab:stats} and \autoref{fig:main_multi}. Notably, although the structure similarity between DKG and GKG is very low (e.g., only 9.8\% for \openalexdbpedia), \ourmodel still performs well, whereas the GNN-based \GraphMambaF remains around 0.50 ACC. This suggests that logical consistency matters more than structural isomorphism in cross-domain knowledge fusion. Overall, \ourmodel improves ACC and F1 by up to 9.5\% over the strongest baselines, while the low absolute F1 of the best baselines on some settings also demonstrates the research value and inherent difficulty of the new DKGF task, which requires explicit modeling of domain relevance and cross-domain granularity gaps.

\subsubsection{Transferability Exploration (RQ1)} To verify that \ourmodel learns transferable logic rather than memorizing entities, we evaluated performance on unseen entities. Specifically, we partition the test set facts into two categories: \emph{seen} and \emph{unseen}. If the head or tail entity in a fact is not present in the training set, it is regarded as unseen. In this setting, a subset of entities does not participate in the training process, making it difficult for the model to locate these truly relevant entities, which makes the test process more challenging. As shown in \autoref{tb:transferability_wiki}, \ourmodel consistently outperforms baselines on unseen entities and remains stable across S1 and S2, whereas \ChatEAF degrades notably when entities are absent from training. 
This indicates that neuro-symbolic execution generalizes to new geometric regions of the DKG. 
We further examine four additional domains: \ourmodel achieves unseen F1 of 0.642 and 0.595 on \drugbankwiki and \umlswiki, and remains above 0.582 on \openalexdbpedia and \crunchbasedbpedia, clearly surpassing \ChatEAF and \TransHF that rely on seen entities or local topology. 
These results show that \ourmodel captures domain-invariant logic rather than entity-specific patterns. 
The full \icewsyago results and the multi-domain transferability results are provided in Appendix~\ref{app:more_Main}.

\begin{table}[t]
\centering
\caption{Transferability and resource robustness on \icewswiki. We report performance on \emph{Seen} and \emph{Unseen} entities across different resource splits. ``\emph{Avg.}'' denotes the average performance across S1 and S2.}
\label{tb:transferability_wiki}
\resizebox{0.85\linewidth}{!}{
\begin{tabular}{l|c|cc|>{\columncolor{lightgray!50}}c>{\columncolor{lightgray!50}}c|cc}
    \toprule
    \multirow{2}{*}{\textbf{Models}} & \multirow{2}{*}{\textbf{Split}} & \multicolumn{2}{c|}{\textbf{Seen}} & \multicolumn{2}{c|}{\cellcolor{lightgray!50}\textbf{Unseen}} & \multicolumn{2}{c}{\textbf{All}} \cr
    \cmidrule(lr){3-4}\cmidrule(lr){5-6}\cmidrule(lr){7-8}
    & & ACC & F1 & ACC & F1 & ACC & F1 \cr
    \midrule
    \multirow{3}{*}{\DistMultF} 
    & S1 & 0.563 & 0.623 & 0.491 & 0.340 & 0.554 & 0.585 \cr
    & S2 & 0.548 & 0.602 & 0.485 & 0.321 & 0.536 & 0.570 \cr
    & \textit{Avg.} & \textit{0.556} & \textit{0.613} & \textit{0.488} & \textit{0.331} & \textit{0.545} & \textit{0.578} \cr
    \midrule
    \multirow{3}{*}{\SelfRAGF} 
    & S1 & 0.565 & 0.570 & 0.594 & 0.372 & 0.576 & 0.544 \cr
    & S2 & 0.552 & 0.565 & 0.581 & 0.366 & 0.544 & 0.545 \cr
    & \textit{Avg.} & \textit{0.559} & \textit{0.568} & \textit{0.588} & \textit{0.369} & \textit{0.560} & \textit{0.545} \cr
    \midrule
    \multirow{3}{*}{\ChatEAF} 
    & S1 & 0.615 & 0.610 & 0.675 & 0.429 & 0.632 & 0.589 \cr
    & S2 & 0.601 & 0.598 & 0.654 & 0.412 & 0.592 & 0.554 \cr
    & \textit{Avg.} & \textit{0.608} & \textit{0.604} & \textit{0.665} & \textit{0.421} & \textit{0.612} & \textit{0.572} \cr
    \midrule
    \multirow{3}{*}{\ourmodel (Ours)} 
    & S1 & \textbf{0.672} & \textbf{0.709} & \textbf{0.731} & \textbf{0.565} & \textbf{0.680} & \textbf{0.690} \cr
    & S2 & \textbf{0.654} & \textbf{0.682} & \textbf{0.712} & \textbf{0.553} & \textbf{0.655} & \textbf{0.654} \cr
    & \textit{Avg.} & \textbf{\textit{0.663}} & \textbf{\textit{0.696}} & \textbf{\textit{0.722}} & \textbf{\textit{0.559}} & \textbf{\textit{0.668}} & \textbf{\textit{0.672}} \cr
    \bottomrule
\end{tabular}
}
\end{table}

\subsubsection{Evaluation of Task Value and Dataset Challenge (RQ2).} Furthermore, we perform a comprehensive analysis of the benchmark datasets, focusing on their complexity of fusion and practical utility. As shown in~\autoref{tb:main_results_general},~\autoref{tb:main_results_crosstask}, and ~\autoref{fig:main_multi} the results indicate a significant potential for improvement in overall performance, as evidenced by the F1 scores of the \sota benchmark configurations, which are only 0.607 and 0.615 for \icewswiki-S2 and \icewsyago-S2, respectively. It is worth noting that the performance of most models in related research tasks, such as entity alignment and knowledge graph completion, exceeds 0.9 in many cases \cite{chatea, kgcmm24}. These findings highlight the inherent difficulty of the task and suggest that there are still numerous unresolved challenges.

% \begin{figure}[t]
% 	\centering
%     \includegraphics[width=1\linewidth]{leida.pdf}
%     \vspace{-10pt}
% 	\caption{The results of the transferability experiment on the \icewswiki and \icewsyago datasets at different training rates. The test data is split into seen (S) and unseen (U) parts based on whether the entity appeared during training. The total results for all (A) test data are also reported. Accuracy (ACC) and F1-score (F1) are shown in the radar charts.}
% 	\label{fig:leida}
% \end{figure}

\begin{table}[t]
\centering
\caption{Ablation study and design verification on \textbf{\icewswiki-S1}. ``\emph{Avg.}'': average of ACC and F1. ``\emph{$\Delta$}'': relative performance drop in Avg.}
\label{tb:ablation}
% \vspace{-8pt}
\setlength{\tabcolsep}{3.5pt}
\resizebox{0.8\linewidth}{!}{
\begin{tabular}{l|ccc|c}
\toprule
\multirow{2}{*}{\textbf{Variant}} & \multicolumn{3}{c|}{\textbf{\icewswiki-S1}} & \multirow{2}{*}{\textbf{$\Delta$}} \cr
\cmidrule(lr){2-4}
& ACC & F1 & Avg. & \cr
\midrule
\rowcolor{gray!10} \textbf{ExeFuse} (Full Model) & \textbf{0.680} & \textbf{0.690} & \textbf{0.685} & - \cr
\midrule
\multicolumn{5}{l}{\textit{\textbf{I. Complexity Validation} (Is the complex pipeline necessary?)}} \\
w/o Rules $\mathcal{L}$ (Single Affine Projection) & 0.605 & 0.612 & 0.609 & -11.1\% \cr
w/o Reasoning (Direct Classification) & 0.577 & 0.583 & 0.580 & -15.3\% \cr
\midrule
\multicolumn{5}{l}{\textit{\textbf{II. Module Necessity} (Do all components contribute?)}} \\
w/o Structure-Aware Encoder ($\phi_{enc}$) & 0.570 & 0.582 & 0.576 & -15.9\% \cr
w/o Neural Rule Selector ($p(R|q^g)$) & 0.620 & 0.623 & 0.622 & -9.2\% \cr
w/o Granularity Calibration ($S_{exec}$) & 0.595 & 0.601 & 0.598 & -12.7\% \cr
\midrule
\multicolumn{5}{l}{\textit{\textbf{III. Design Choice Verification} (Are choices optimal?)}} \\
Replace \textit{Affine} Rules with \textit{MLP} & 0.665 & 0.672 & 0.669 & -2.3\% \cr
Replace \textit{KDE} with \textit{Min-Distance} & 0.648 & 0.655 & 0.652 & -4.8\% \cr
\bottomrule
\end{tabular}
}
% \vspace{-10pt}
\end{table}

\begin{table}[t]
\centering
\caption{Accuracy and cost analysis of \ourmodel and benchmark configurations on \icewswiki-S1. \emph{Avg.tokens}/\emph{Avg.time}: the average tokens/time cost of the model.}%都是运行5次的平均时间
% \vspace{-8pt}
\setlength{\tabcolsep}{3pt}
\resizebox{0.8\linewidth}{!}{
\begin{tabular}{l|ccc|cc}
    \toprule
    \multirow{2}{*}{\ \ \ \ \ \textbf{Settings}} &\multicolumn{3}{c}{\icewswiki-S1 (Accuracy)} &\multicolumn{2}{|c}{Cost (Efficiency and Effectiveness)}\cr
    \cmidrule(lr){2-4}\cmidrule(lr){5-6}
    & ACC & F1 & \emph{Avg.} & \emph{Avg. tokens} & \emph{Avg. time (seconds)}\cr
    \midrule 
    
    \SelfConsistencyF & 0.592 & 0.561& 0.577& 41,561,586& 178,041.6\cr   

    \ChatEAF & 0.596& 0.566& 0.581 & 48,162,276& 254,417.4\cr   

    \TransHF & \underline{0.640}& \underline{0.591}& \underline{0.583} & \textbf{0} & \underline{55.7}\cr
    
    \SimpleHHEAF & 0.490& 0.499& 0.495 & \textbf{0} & 2628.2\cr

    \midrule
    \ourmodel (Ours) & \textbf{0.680}& \textbf{0.690}& \textbf{0.685} & \textbf{0}& \textbf{42.6}\cr
    \bottomrule
\end{tabular}}
\label{tb: compare_cost}
\vspace{-11pt}
\end{table}

\subsection{Ablation Study (RQ3)}\label{exp:abl}
To rigorously evaluate the necessity of our architecture, we conducted a comprehensive ablation study on \icewswiki-S1 (see \autoref{tb:ablation}). Specifically, we verify how each component contributes to addressing the two core DKGF challenges.

\subsubsection{Justification of Complexity vs. Performance}
We first address the concern of whether the neuro-symbolic complexity is warranted. \textcircled{1} \textbf{Single Affine Projection.} We replaced the multi-rule logic execution with a single learnable linear transformation matrix. The performance drops significantly ($\Delta$-11.1\%) to an F1 of 0.612. Notably, this result is comparable to the translation-based baseline \TransEF (F1 0.630) in~\autoref{tb:main_results_general}. This confirms that simple linear mappings hit a performance bottleneck and are insufficient to bridge the complex semantic gap between GKG and DKG. \textcircled{2} \textbf{Direct Classification.} Removing the reasoning module entirely leads to the worst performance ($\Delta$-15.3\%), proving that the encoder alone cannot resolve the ambiguity of relevance without explicit logical transformation.

\subsubsection{Impact of Key Modules \& Challenge Resolution}
We validate the contributions of each stage in the \textit{Fact-as-Program} paradigm and their roles in solving specific tasks. \textcircled{1} \textbf{Latent Program Induction:} Replacing the Structure-Aware Predicate ($\phi_{enc}$) with a standard linear encoder leads to the largest drop ($\Delta$-15.9\%). This confirms that treating facts as structured programs, preserving the compositional logic of $(h,r,t)$, is the foundational basis for executable reasoning. \textcircled{2} \textbf{Neuro-Symbolic Execution (Addressing Challenge I).} Removing the dynamic rule selector (using uniform weights) degrades performance by 9.2\%, while removing execution entirely causes a 15.3\% drop. This demonstrates that static similarity matching fails to capture implicit logical connections. The proposed neuro-symbolic execution effectively resolves ambiguity of domain relevance by inferring logical reachability beyond surface text. \textcircled{3} \textbf{Target Space Grounding (Addressing Challenge II).} Removing the granularity calibration mechanism ($S_{exec}$) leads to a 12.7\% drop. This finding is critical: it indicates that even if a fact is logically relevant, it may still fail to fuse correctly due to granularity misalignment. The grounding mechanism acts as a necessary compiler check, ensuring the transformed knowledge aligns with the specific density and granularity of the target domain.

\subsubsection{Verification of Design Choices}
Finally, we justify specific neuro-symbolic reasoning technical decisions. \textcircled{1} \textbf{Affine vs. MLP Rules.} Replacing affine operators with non-linear MLPs degrades performance by 2.3\%. We attribute this to the higher parameter count of MLPs leading to overfitting on the sparse DKG data, whereas affine transformations offer a robust and parameter-efficient representation for logic rules. \textcircled{2} \textbf{KDE vs. Min-Distance.} Replacing KDE-based density estimation with simple Euclidean distance to the nearest centroid causes a 4.8\% drop. This confirms that density-based verification is more reliable for quantifying granularity alignment than point-to-point distance metrics.

%待优化
\subsection{Efficiency Analysis (RQ4)}\label{exp:efficiency}%半小时
We evaluate \ourmodel's efficiency against two categories: \textcircled{1} lightweight general-purpose and cross-task adaptation configurations (e.g., \TransHF, \SimpleHHEAF), which are efficient but limited in capturing complex domain interactions, and \textcircled{2} LLM-based general-purpose and cross-task adaptation configurations (e.g., \ChatEAF, \SelfConsistencyF), which exploit LLMs but incur high computational overhead. As shown in~\autoref{tb: compare_cost} , \ourmodel achieves the best trade-off between effectiveness and efficiency: it delivers the highest accuracy (Avg. 0.685) while eliminating token consumption through our small-model paradigm, requiring only \textbf{42.6 seconds} with substantial performance gains (17.9\%-38.4\%). This validates the advantage of our \textit{Fact-as-Program} paradigm, which provides a new perspective for DKGF by reformulating knowledge fusion as executable neuro-symbolic reasoning without relying on expensive LLM API calls. These results validate its robustness in overcoming performance bottlenecks on the DKGF benchmark. The further analysis of theoretical complexity can be found in Appendix~\ref{sec:complexity}.

\section{Conclusion and Future Work}
This paper introduces domain-specific knowledge graph fusion (DKGF), a new task aimed at enriching specialized knowledge graphs with general knowledge while overcoming the challenges of relevance ambiguity and granularity misalignment. To address these issues, we propose \ourmodel, a neuro-symbolic framework based on the \emph{Fact-as-Program} paradigm. \ourmodel treats knowledge fusion as an executable logical process, employing neuro-symbolic execution to infer deep relevance and a target space grounding mechanism to calibrate knowledge granularity. We also establish the standardized evaluation suite with six benchmark datasets. Extensive experiments demonstrate that \ourmodel effectively bridges domain barriers to achieve superior fusion performance. In the future, we plan to enhance the framework by integrating LLMs to further improve the interpretability and reasoning depth of the neuro-symbolic execution process.

\bibliographystyle{ACM-Reference-Format}
\bibliography{ref}

\appendix

\section*{Appendix}

\section{More Method Details}

\subsection{Complexity and Scalability Analysis}
\label{sec:complexity}
We analyze the inference complexity to demonstrate the scalability of \ourmodel. Let $|F^g|$ be the number of GKG facts, $d_{emb}$ be the embedding dimension, and $K$ be the number of prototypes. The complexity per fact consists of three parts: \textcircled{1} \textbf{Latent Program Induction.} $O(d_{emb}^2)$ for the structure-aware neural predicate, dominated by matrix-vector multiplications. \textcircled{2} \textbf{Neuro-Symbolic Execution.} $O(|\mathcal{L}| \cdot d_{emb})$ for attention calculation and element-wise affine transformations. Since $|\mathcal{L}|$ is small (typically $<100$), this step is highly efficient. \textcircled{3} \textbf{Target Space Grounding:} $O(K \cdot d_{emb})$ for measuring the distance against $K$ prototypes in the Target State Space. Using clustered prototypes ($K \ll |F^d|$) avoids the prohibitive cost of traversing the entire DKG. Consequently, the overall complexity is \textbf{$O(|F^g| \cdot (d_{emb}^2 + |\mathcal{L}|d_{emb} + Kd_{emb}))$}. This scales linearly with the GKG size $|F^g|$, offering a significant efficiency advantage over traditional pair-wise matching methods that typically require quadratic $O(|F^g| \cdot |F^d|)$ operations.

\subsection{Theoretical Analysis}

\label{sec:theory}
To justify the design choices of \textsc{ExeFuse}, we analyze the expressive power of the neuro-symbolic mapping and the geometric interpretation of the grounding mechanism. Detailed derivations are provided in our anonymous repository\footnote{\url{https://anonymous.4open.science/r/DKGF_1-D85B/}\label{repository}}.

\begin{proposition}[Expressive Completeness of Neuro-Symbolic Mapping]
\label{prop:expressiveness}
Let $\mathcal{T}: \mathcal{Q} \to \mathcal{Q}$ be the ideal transition function mapping source facts to the target domain. The affine rule operators $\mathcal{L}$, combined with the non-linear neural selector $\mathcal{P}$, constitute a universal approximator. Specifically, the expected execution $\hat{q}^d = \sum_{R \in \mathcal{L}} p(R|q^g) \cdot (w_R \odot q^g + b_R)$ can approximate any continuous function on a compact set to arbitrary precision, given sufficient rules $|\mathcal{L}|$.
\end{proposition}

\textit{Analysis.} Proposition~\ref{prop:expressiveness} justifies the use of multiple discrete logic rules instead of a single projection matrix. While a single affine transformation is limited to linear mappings, our mechanism functions as a \textit{mixture of affine experts}. The neural selector dynamically weights these experts, enabling the model to fit complex, non-linear decision boundaries required to bridge the semantic gap between GKG and DKG, ensuring that valid domain states are reachable within the model's hypothesis space.

\begin{proposition}[Granularity Alignment via Density Maximization]
\label{prop:density}
Let $p_{\text{data}}(x)$ be the probability density function of valid facts in the target DKG space $\mathcal{Q}^d$. The executability score $S_{exec}(f^g)$, defined via kernel density estimation over prototypes, acts as a proxy for the log-likelihood of the fused fact under the target distribution. Maximizing $S_{exec}(f^g)$ is equivalent to minimizing the divergence between the generated fact embedding and the underlying data manifold of $G^d$.
\end{proposition}

\textit{Analysis.} Proposition~\ref{prop:density} clarifies that $S_{exec}$ is not merely a distance heuristic but a density-based verification mechanism. By optimizing this score, \textsc{ExeFuse} explicitly constrains the fused facts to reside in high-density regions of the target space (i.e., the valid semantic manifold of the domain). This theoretically guarantees that the generated facts are not just logically relevant but also conform to the specific granularity and structural patterns characteristic of the DKG.

\section{More Experiments}
\label{app:Experiments}

\subsection{DKGF Benchmarks} 
Currently, there is no benchmark that effectively studies and evaluates the DKGF task. To this end, we construct six new DKGF benchmark datasets across a diverse set of critical fields (Political, Biomedical, Academic, and Business) to support comprehensive knowledge integration and cross-domain evaluation. As shown in~\autoref{tab:stats}, the DKGs are drawn from highly specialized databases: ICEWS provides data on temporal political events capturing evolving geopolitical relationships; within the biomedical domain, DrugBank~\cite{DrugBank} offers rich biochemical and pharmacological knowledge critical for drug discovery, while UMLS \cite{UMLS} encapsulates complex medical and clinical concepts; OpenAlex~\cite{OpenAlex} catalogs a massive scale of academic entities and scientific literature; and Crunchbase \cite{Crunchbase} structures business, investment, and financial entities. 

General knowledge is sourced from Wikidata~\cite{DBLP:conf/semweb/ErxlebenGKMV14}, YAGO~\cite{YAGO}, and DBpedia~\cite{DBpedia}, all of which are high-quality, large-coverage resources that provide rich background facts and entity attributes. During dataset construction, we align and fuse cross-domain knowledge by incorporating highly relevant information from these comprehensive GKGs into the respective DKGs. This process mitigates the inherent incompleteness of specialized domain knowledge, improves multidimensional entity understanding, and offers robust data support for downstream knowledge reasoning and applications across multiple disciplines. Due to space limitations, further construction details and data are available in our anonymous repository\footnote{\url{https://anonymous.4open.science/r/DKGF_1-D85B/}\label{repository}}.

\subsection{Benchmark Suite Construction \& Adaptation} 
Since DKGF is a nascent task with no pre-existing specialized solutions, a primary contribution of this work is to establish a comprehensive and standardized evaluation suite to facilitate future research. These configurations are categorized as follows: \textcircled{1} \textbf{General-purpose Configurations.} To establish benchmark performance using universal architectures, this category employs broadly applicable methods (e.g., TransE, GNN, BERT, LLMs) rather than task-specific designs. This category of configurations is mainly based on current advanced and classic general methods, including rule-based methods (i.e., ``Rule.''), such as \StringMatch~\cite{aaaistringmatch} and \TFIDF~\cite{VLDBTFIDF}; Translation-based methods (i.e., ``Trans.'') such as \TransEF~\cite{transE}, \TransHF~\cite{DBLP:conf/iclr/RotatE}, \DistMultF~\cite{DBLP:journals/corr/DistMult}, and \ComplExF~\cite{DBLP:conf/icml/ComplEx}; GNN-based methods (i.e., ``GNN'') like \GCNF~\cite{GCNnips24,GCNKG22}, \TransGNNF~\cite{transgnn}, and \GraphMambaF~\cite{GraphMamba}; Generative methods (i.e., ``Generative.'') like \BERTF~\cite{bert}, \ICLF~\cite{ICLACL25,cot2022}, \SelfConsistencyF~\cite{selfcot,tkde24}, and \SelfRAGF~\cite{selfrag}; \textcircled{2} \textbf{Cross-task Adaptation Configurations.} This category of configuration mainly involves improving representative methods from current related research tasks to adapt to the DKGF task, including entity alignment (i.e., ``EA.''), such as \SimpleHHEAF~\cite{simplehhea} and \ChatEAF~\cite{chatea}; knowledge graph completion (i.e., ``KGC.''), such as \KGBERTF~\cite{KG-BERT,kgcmm24}, \KGLLaMAF~\cite{KG-LLaMA,kgcmm24}, \KoPAF~\cite{kgcmm24} and \PKGC~\cite{PKGC}; and relation triple extraction (i.e., ``RTE.''), like  \NoGenBARTF~\cite{tripleSIGIR}, and \NoGenTfiveF~\cite{tripleSIGIR}. We systematically adapted 21 representative methods from related fields (e.g., KGC, Entity Alignment) to the DKGF setting using a unified \textit{Fusion Scoring Protocol}. Specifically, to bridge the gap between their original tasks and DKGF: (1) for translation and GNN-based methods (e.g., TransE, GCN), we retrained them on the joint graph $G^d \cup G^g$ to score candidate fusion triples $(h^d, r^d, t^g)$ based on link plausibility; (2) for Entity Alignment methods (e.g., ChatEA), we utilized aligned entity pairs to project potential facts from $G^g$ into the $G^d$ semantic space. This rigorous adaptation not only ensures a fair comparison but also provides a robust and reproducible benchmark configuration ecosystem for the community.

\paragraph{Baseline adaptation protocol.}
Adapting existing methods to DKGF is non-trivial, as the task involves cross-graph fusion without shared embedding spaces.
For each baseline, we follow its strongest feasible adaptation under a unified information setting, without introducing task-specific heuristics unavailable to other methods.
For example, translation-based KGC models (e.g., TransE) are trained on the unified graph $G^d \cup G^g$ with disjoint entity embeddings, and candidate triples $(h^d, r^d, t^g)$ are scored using their standard plausibility functions.
Entity alignment methods leverage aligned entity pairs to project candidate facts, while RTE-based methods operate purely on structured triples without access to gold fusion labels.

\begin{table}[t]
\centering
\caption{Transferability and resource robustness on \icewsyago. Metrics are averaged across splits to demonstrate fusion stability under varying resource constraints.}
\label{tb:transferability_yago}
\setlength{\tabcolsep}{3pt} 
\resizebox{0.9\linewidth}{!}{
\begin{tabular}{l|c|cc|>{\columncolor{lightgray!50}}c>{\columncolor{lightgray!50}}c|cc}
    \toprule
    \multirow{2}{*}{\textbf{Models}} & \multirow{2}{*}{\textbf{Split}} & \multicolumn{2}{c|}{\textbf{Seen}} & \multicolumn{2}{c|}{\cellcolor{lightgray!50}\textbf{Unseen}} & \multicolumn{2}{c}{\textbf{All}} \cr
    \cmidrule(lr){3-4}\cmidrule(lr){5-6}\cmidrule(lr){7-8}
    & & ACC & F1 & ACC & F1 & ACC & F1 \cr
    \midrule
    \multirow{3}{*}{\DistMultF} 
    & S1 & 0.531 & 0.564 & 0.472 & 0.315 & 0.525 & 0.536 \cr
    & S2 & 0.582 & 0.591 & 0.490 & 0.342 & 0.575 & 0.562 \cr
    & \textit{Avg.} & \textit{0.557} & \textit{0.578} & \textit{0.481} & \textit{0.329} & \textit{0.550} & \textit{0.549} \cr
    \midrule
    \multirow{3}{*}{\SelfRAGF} 
    & S1 & 0.602 & 0.585 & 0.582 & 0.334 & 0.594 & 0.540 \cr
    & S2 & 0.585 & 0.550 & 0.601 & 0.355 & 0.577 & 0.532 \cr
    & \textit{Avg.} & \textit{0.594} & \textit{0.568} & \textit{0.592} & \textit{0.345} & \textit{0.586} & \textit{0.536} \cr
    \midrule
    \multirow{3}{*}{\ChatEAF} 
    & S1 & 0.621 & 0.592 & 0.688 & 0.431 & 0.649 & 0.576 \cr
    & S2 & 0.602 & 0.584 & 0.662 & 0.401 & 0.592 & 0.541 \cr
    & \textit{Avg.} & \textit{0.612} & \textit{0.588} & \textit{0.675} & \textit{0.416} & \textit{0.621} & \textit{0.559} \cr
    \midrule
    \multirow{3}{*}{\ourmodel (Ours)} 
    & S1 & \textbf{0.650} & \textbf{0.685} & \textbf{0.724} & \textbf{0.560} & \textbf{0.661} & \textbf{0.662} \cr
    & S2 & \textbf{0.641} & \textbf{0.677} & \textbf{0.719} & \textbf{0.548} & \textbf{0.633} & \textbf{0.657} \cr
    & \textit{Avg.} & \textbf{\textit{0.646}} & \textbf{\textit{0.681}} & \textbf{\textit{0.722}} & \textbf{\textit{0.554}} & \textbf{\textit{0.647}} & \textbf{\textit{0.660}} \cr
    \bottomrule
\end{tabular}
}
\end{table}

\begin{figure}[t]
	\centering
    \includegraphics[width=1\linewidth]{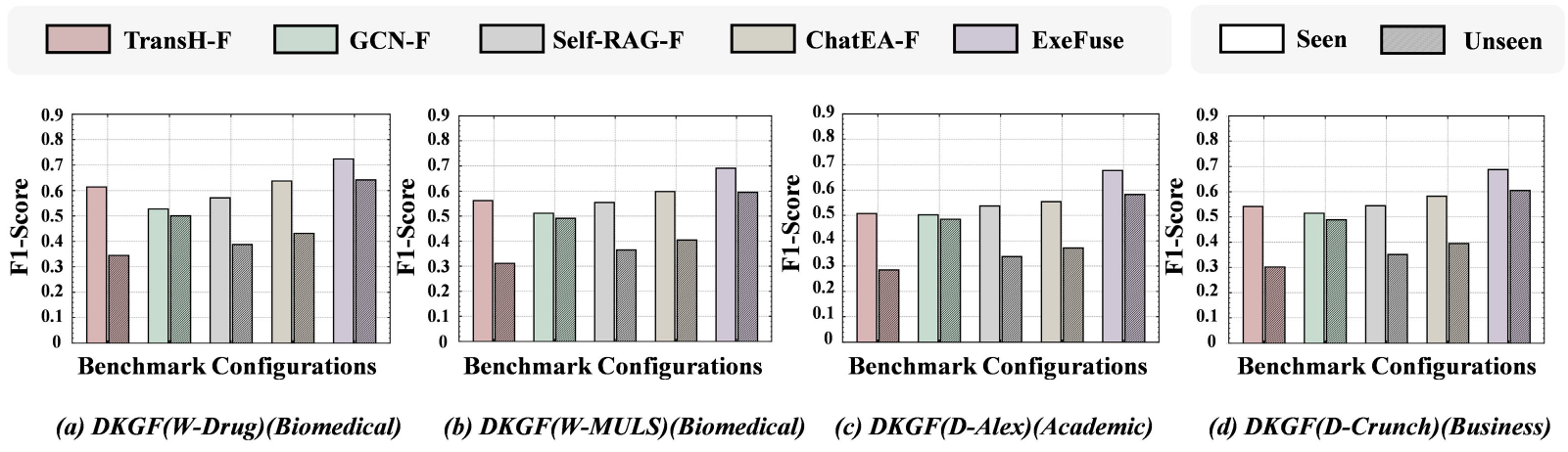}
    \vspace{-15pt}
	\caption{Transferability results (F1) across multiple domains (Biomedical, Academic, and Business).}
	\label{fig:multi_seenunseen}
    \vspace{-10pt}
\end{figure}

\subsection{More Main Results}\label{app:more_Main}
This section provides additional transferability results. 
Table~\ref{tb:transferability_yago} reports the full seen/unseen results on \icewsyago, complementing Table~\ref{tb:transferability_wiki} in the main text. 
Figure~\ref{fig:multi_seenunseen} reports the transferability results (F1) on unseen entities across biomedical, academic, and business domains. 
Together with the main-text results, they confirm that \ourmodel generalizes beyond seen entities.

\begin{figure}[t]
  \centering
   \vspace{-1pt}
\includegraphics[width=0.6\linewidth]{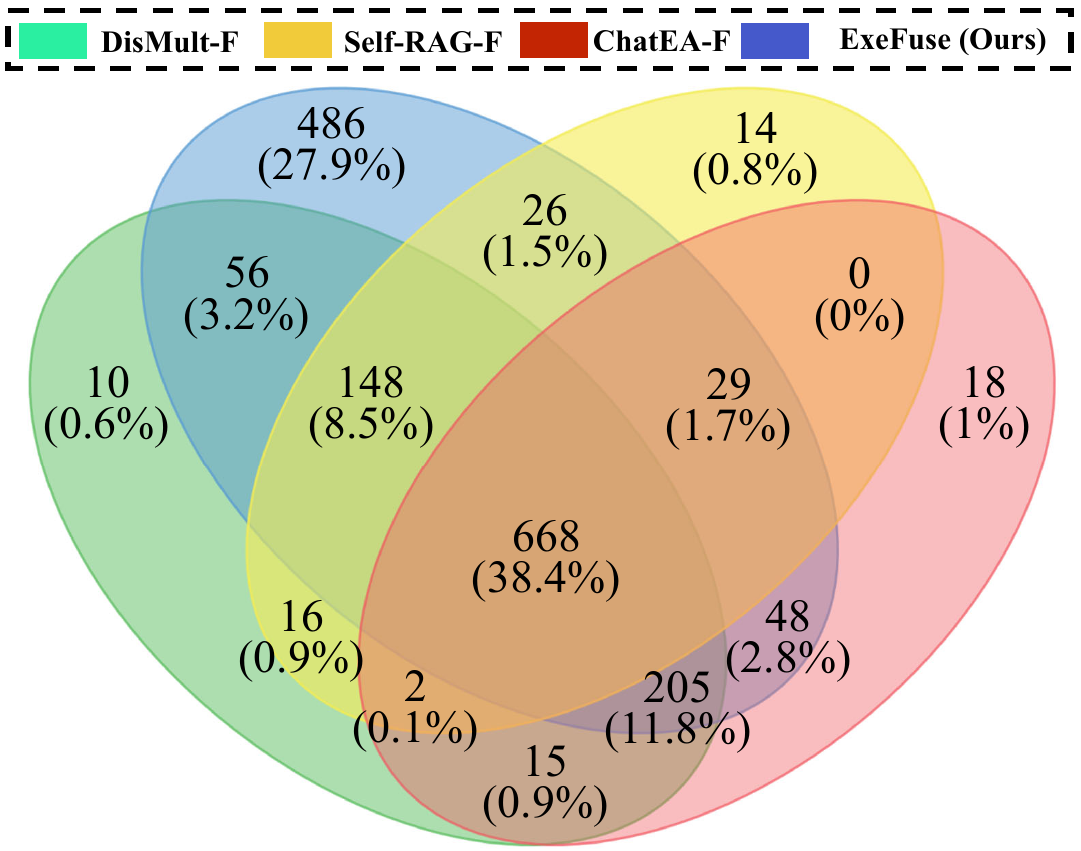}
% \vspace{-12pt}
  \caption{The venn diagram of correct predictions on \icewswiki-S1. Intersections show shared same predictions.}
 \vspace{-10pt}
  \label{case-study}
  % \vspace{-5pt}
\end{figure}

\begin{table}[t]
\centering
\caption{Performance analysis of the new task ``Relevant Entity Finding (REF)''.}
% \vspace{-10pt}
\resizebox{\columnwidth}{!}{
\begin{tabular}{l|>{\columncolor{lightgray!50}}c|cc|>{\columncolor{lightgray!50}}c|>{\columncolor{lightgray!50}}c||>{\columncolor{lightgray!50}}c|cc|>{\columncolor{lightgray!50}}c|>{\columncolor{lightgray!50}}c|}
    \toprule
    \multirow{2}{*}{\ \ \textbf{Models}} 
    &\multicolumn{5}{c||}{\textbf{\icewswiki-S1}} 
    &\multicolumn{5}{c}{\textbf{\icewsyago-S1}} \cr
    \cmidrule(lr){2-6}\cmidrule(lr){7-11}
    & ACC & P & R & F1 & Avg. 
    & ACC & P & R & F1 & Avg.\cr
    \midrule

    \DistMultF 
    & 0.975 & 0.830 & 0.840 & 0.835 & 0.905 
    & 0.946 & 0.853 & 0.730 & 0.787 & 0.866 \cr

    \SelfRAGF 
    & 0.970 & 0.849 & 0.731 & 0.785 & 0.878 
    & 0.936 & 0.897 & 0.603 & 0.721 & 0.829 \cr

    \ChatEAF 
    & 0.973 & 0.888 & 0.737 & 0.806 & 0.889 
    & 0.938 & 0.937 & 0.583 & 0.719 & 0.828 \cr

    \ourmodel 
    & \textbf{0.982} & 0.968 & 0.791 & \textbf{0.871} & \textbf{0.927} 
    & \textbf{0.958} & 0.958 & 0.721 & \textbf{0.823} & \textbf{0.890} \cr

    \bottomrule
\end{tabular}}
\label{tb:entity_finding}
% \vspace{-10pt}
\end{table}

\subsection{Relevant Entity Finding}\label{exp:more}
In the DKGF process, accurate identification of general knowledge relevant to the DKG is a critical prerequisite to achieve reliable fusion. Failure to discover the correct cross-domain related entities results in erroneous fact generation and downstream inference.We formulate and systematically study an essential yet under-explored task in the DKGF pipeline, namely Relevant Entity Finding (REF), which aims to identify a set of entities in the GKG that are semantically related to the DKG. We compare \ourmodel against embedding-based (\DistMultF) and LLM-based (\SelfRAGF, \ChatEAF) benchmark configurations. As shown in \autoref{tb:entity_finding}, \ourmodel consistently outperforms all benchmarks and achieves an F1 of \textbf{0.871} on \icewswiki-S1, significantly surpassing \ChatEAF (0.806). This is because embedding methods rely on surface similarity that fails for implicit connections, while LLMs lack structural constraints and may hallucinate relevance; in contrast, \ourmodel defines relevance via the Fact-as-Program paradigm and \textbf{Logical Reachability} (Sections~\ref{sec:overview} and~\ref{sec:execution}), treating an entity as relevant only if its associated fact can be transformed into a valid state in the DKG's Target State Space via neuro-symbolic execution, thereby filtering ambiguous noise. Moreover, the REF trends closely mirror the main DKGF results, empirically validating that resolving domain-relevance ambiguity (Stage 2) is a fundamental prerequisite for effective knowledge fusion.

\subsection{Case study}
\label{exp:case}
To intuitively demonstrate \ourmodel's effectiveness, we conduct a multi-level analysis combining macro-statistics and micro-cases.
\textbf{(1) Macro Analysis:} The Venn diagram in \autoref{case-study} shows that \ourmodel correctly identifies a unique subset of fused facts ($27.9\%$) missed by both embedding-based and LLM-based baselines. This superiority stems from our neuro-symbolic approach, which captures complex logical dependencies and granularity alignments inaccessible to methods relying solely on surface text or static embeddings. 
\textbf{(2) Micro Analysis:} Consider the challenging inference: \emph{(Turki bin Faisal Al Saud, Host a visit, Brookings Institution)}. While \DistMultF fails due to low surface similarity, and LLMs (e.g., \ChatEAF) fail to bridge the semantic gap between the retrieved GKG fact \emph{(employer)} and the target DKG relation \emph{(Host a visit)}, \ourmodel succeeds by treating the GKG fact as an executable program. Our \textit{Neuro-Symbolic Execution} identifies the latent logic bridging these relations, while \textit{Target Space Grounding} ensures the result aligns with the fine-grained granularity of political events. This confirms \ourmodel's ability to resolve ambiguity through logical reachability. Due to space constraints, we provide additional diverse case studies and visualizations in our anonymous repository\footref{repository}.

\end{document}